\documentclass[letterpaper]{article} 
\usepackage{aaai23-r2hcai}  
\usepackage{times}  
\usepackage{helvet}  
\usepackage{courier}  
\usepackage[hyphens]{url}  
\usepackage{graphicx} 
\def\UrlFont{\rm}  
\usepackage{natbib}  
\usepackage{caption} 
\usepackage{xcolor}
\usepackage{algorithm}
\usepackage{algorithmic}
\usepackage{subcaption}
\usepackage{graphicx}
\usepackage{mathtools}
\usepackage{amsmath,amssymb}
\usepackage{breqn}
\usepackage{physics}

\DeclareMathOperator{\Rh}{\mathcal{\tilde{R}}^h}

\newcommand{\Rpsi}{R^{h}_{\psi}}

\usepackage{newfloat}
\usepackage{listings}
\DeclareCaptionStyle{ruled}{labelfont=normalfont,labelsep=colon,strut=off} 
\floatstyle{ruled}
\newfloat{listing}{tb}{lst}{}
\floatname{listing}{Listing}
\usepackage{fancyhdr}
\fancypagestyle{firstpagehf}
{
   \fancyhf{}
   \fancyhead[HC]{The AAAI 2023 Workshop on Representation Learning for Responsible Human-Centric AI (R$^2$HCAI)}
   \fancyfoot[C]{\thepage}
}

\title{Data Driven Reward Initialization for Preference based Reinforcement Learning}
\author{
   Mudit Verma\textsuperscript{\rm 1} \quad
   Subbarao Kambhampati \textsuperscript{\rm 1} \quad
}
\affiliations{
    \textsuperscript{\rm 1} SCAI, Arizona State University\\
    muditverma@asu.edu, rao@asu.edu
    
}

\begin{document}
\thispagestyle{firstpagehf}
\maketitle

\begin{abstract}
Preference-based Reinforcement Learning (PbRL) methods utilize binary feedback from the human in the loop (HiL) over queried trajectory pairs to learn a reward model in an attempt to approximate the human's underlying reward function capturing their preferences. In this work, we investigate the issue of a high degree of variability in the initialized reward models which are sensitive to random seeds of the experiment. This further compounds the issue of degenerate reward functions PbRL methods already suffer from. We propose a data-driven reward initialization method that does not add any additional cost to the human in the loop and negligible cost to the PbRL agent and show that doing so ensures that the predicted rewards of the initialized reward model are uniform in the state space and this reduces the variability in the performance of the method across multiple runs and is shown to improve the overall performance compared to other initialization methods.

\end{abstract}

\section{Introduction}

Reinforcement Learning (RL) allows AI agents to learn by trial and error \cite{mnih2015human, arulkumaran2017deep, verma2019making}. While the dynamics of the interaction are governed by the environment in which the agent executes a chosen action, the only task-relevant learning signal comes from the given reward function. Therefore, several of the recent works on RL \cite{ird, reward-hacking-example} have pointed out the importance of the specified reward function in achieving a given task. Hence, despite recent successes of Deep Reinforcement Learning for complex high dimensional state and action space domains, prior works have found that specifying reward functions for seemingly ``easy" tasks could be extremely hard, to begin with, and this becomes even more challenging with potential issues of reward hacking \cite{reward-hacking-example, reward-hacking-init} that may impact Human-AI trust \cite{zahedi2021trust, zahedi2022modeling}.  In fact, works in explainable AI literature have attempted at explaining environment dynamics to human in the loop \cite{blackbox, adv-conf, gopalakrishnan2021computing, gopalakrishnan2021synthesizing}, indicating that requesting detailed reward specifications over low-level state space is challenging for HiL \cite{verma2021perfect}.

Several solutions to the reward learning problem have been proposed in the past like learning from demonstration (LfD) \cite{lfd} or advice \cite{expand, guan2020explanation}, imitation learning \cite{imitation_learning} and preference-based reinforcement learning \cite{prior, pebble, surf, christiano}. While each paradigm has its own challenges, Preference-based Reinforcement Learning (PbRL) relies on binary feedback of a human in the loop (HiL) on queried trajectory pairs to convey what the user ``wants" or ``prefers", in contrast to a HiL performing a demonstration for imitation learning or LfD line works which could be prohibitively expensive, or human may not have expertise on specific action choices, for example, a human may find it easier to provide their preference over two trajectories regarding which one appears more as a humanoid back-flip than to demonstrate it.

\begin{figure}[h]
    \centering
    \includegraphics[width=0.45\textwidth]{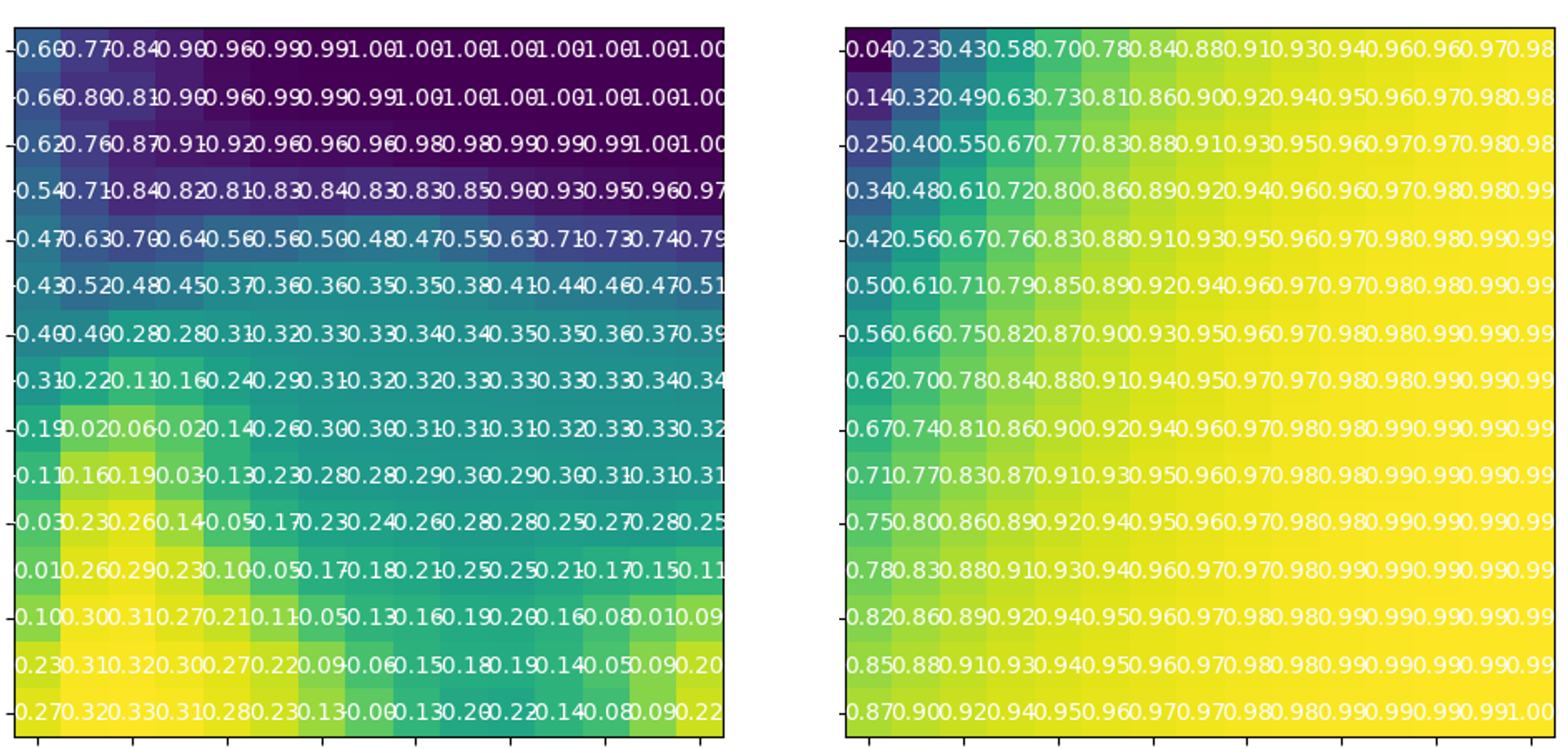}
    \caption{``Patchy" or non-uniform reward prediction by using Kaiming-Uniform weight initialization (left), and the predicted reward values of states by the proposed data-driven reward initialization method over a 15x15 gridworld. Numbers in each cell show the maximum reward,$\Rh$, that the agent upon taking an action in that cell.}
    \label{fig:15x15_grid}
\end{figure}

In Preference-Based Learning methods, recent works \cite{pebble, surf, adv-conf, christiano, prior, soni2022towards} have looked at learning the underlying human reward function from human binary feedbacks on trajectory pairs queried by the RL agent. In this work, we focus our attention to the class of PbRL works that attempt to learn (or approximate) a reward model from human preferences.  We first discuss that learning a reward model from a finite, possibly small number of trajectory preference queries can potentially lead to degenerate solutions (that is, significantly different, multiple reward models can equally explain the given trajectory preferences, thereby yielding very different agent policies). Now, although this is majorly caused by the limited feedback queries we found that high variability in the learned reward model and subsequently the learned agent policy is also seen when we vary the reward function initialization strategy. We find that the PbRL reward learning process is extremely sensitive to the reward function initialization and changing the random seed for the initialization can produce extremely positive or negative results. In this view, we propose a data-driven reward function initialization technique (inspired from data-driven weight initialization methods \cite{weight_init_techniques}) that not only stabilizes the resultant performance across multiple runs but also improves the agent's performance at learning the underlying reward function at potentially zero additional cost to the HiL or the agent being trained.

\section{Background}

Preference-based Reinforcement learning agents sample pairs of trajectories $\tau_0, \tau_1$ using a random policy or via the policy $\pi_\phi$ being learned. An image projection of these trajectory pairs is shown to the human (with image space as the established lingua franca \cite{lingua-franca} between the agent and HiL) in the loop where they provide binary feedback to highlight their relative preference of one trajectory over the other, hence populating a buffer $D_\tau$ with tuples $<\tau_0^i, \tau_1^i, y^i>$. Extending from a typical reinforcement learning setup, preference-based RL agents solve the reward learning problem, ${P} = <\mathcal{M}\setminus r, \Rpsi, \pi^H, Q>$ \cite{shah2020benefits} where the agent models the world as a Markov Decision Process $\mathcal{M}\setminus r = <S, T, A>$ without any environment reward, and instead uses a learned $\Rpsi$ reward model by querying $q \in Q$ questions of the form of trajectory pairs to the HiL to obtain binary feedbacks regarding their preference. Recent PbRL works have resorted to supervised learning to solve this problem of reward learning. First, they approximate the underlying human reward model $\Rh$ by a bounded, parameterized function approximator $\Rpsi : S \rightarrow \mathbb{R}$. They use the Bradley Terry model \cite{bradley-terry} to obtain the probability of one trajectory being preferred over another, $P_\psi$, as the softmax \cite{bishop} over predicted returns (as the sum of predicted rewards) over the two trajectories. 
 \begin{equation}
     P_{\psi}[\tau_0 \succ \tau_1] = \frac{\exp(\sum_{t}R_h(s_t^0, a_t^0))}{\sum_{i \in \{0,1\}} \exp(\sum_{t}R_h(s_t^i, a_t^i))}
 \end{equation}
This is essentially a classifier to approximate human preference feedbacks which can be trained by minimizing the cross entropy loss between the predictions and the ground truth human labels as follows : 
\begin{dmath}
    \mathcal{L}_{CE}=-\displaystyle \mathop{\mathbb{E}}_{(\tau_0, \tau_1, y)\sim \mathcal{D}}[y(0)\text{log}P_\psi[\tau_0 \succ \tau_1] + {y(1)\text{log}P_\psi[\tau_1 \succ \tau_0]]}
\end{dmath}

Several works have extended the above approach in various ways, by building priors over the reward space, improving the query method, improving the exploration strategy, and state or temporal augmentation to name a few. In this work, we are interested in realizing how susceptible is the backbone algorithm used in several PbRL works to reward initialization and subsequently propose a simple initialization method that poses zero cost to the human in the loop as well as to the PbRL agent.



\section{Problem}

It is well known in the Inverse Reinforcement Learning community that reward model learning suffers from the issue of degeneracy. We discuss reward model degeneracy in the context of preference-based reinforcement learning. Even with a large amount of feedback queries to the human in the loop to elicit their preferences, as pointed out in \cite{ng2000algorithms}, the base case of $\Rpsi(s) = 0 \; \forall s \in S$ will always be a solution as any policy $\pi$, including a random policy as well as the optimal human policy $\pi_h^*$ according to their underlying reward model $\Rh$, is optimal. Without a good inductive bias, we posit that reducing the number of queries to the human in the loop will eventually allow for more degenerate solutions. 

However, recent works \cite{pebble, surf, christiano, prior} all suffer from the issue of degeneracy more than one may expect. In our experiments with \cite{pebble} baseline, we found that the method has high variance in the performance of the learned reward model and is very sensitive to extraneous parameters like random seen of the experiment. Now, although several other sources contributing towards the reward function degeneracy exist and may even be unidentified, we found that reward initialization can have a big impact on the subsequent agent's reward recovery and policy performance.

\subsection{Sensitivity to Reward Initialization}
\label{sec:sensitivity}
Typically, RL (and PbRL) works do not focus on reward function initialization and try to subdue its effects on the final performance by showcasing mean performance over multiple experiments runs. However, in our initial investigation, we found that a change in the initial reward estimates can have a substantial impact on the final performance.

\begin{figure*}[h]
    \centering
    \includegraphics[width=0.8\textwidth]{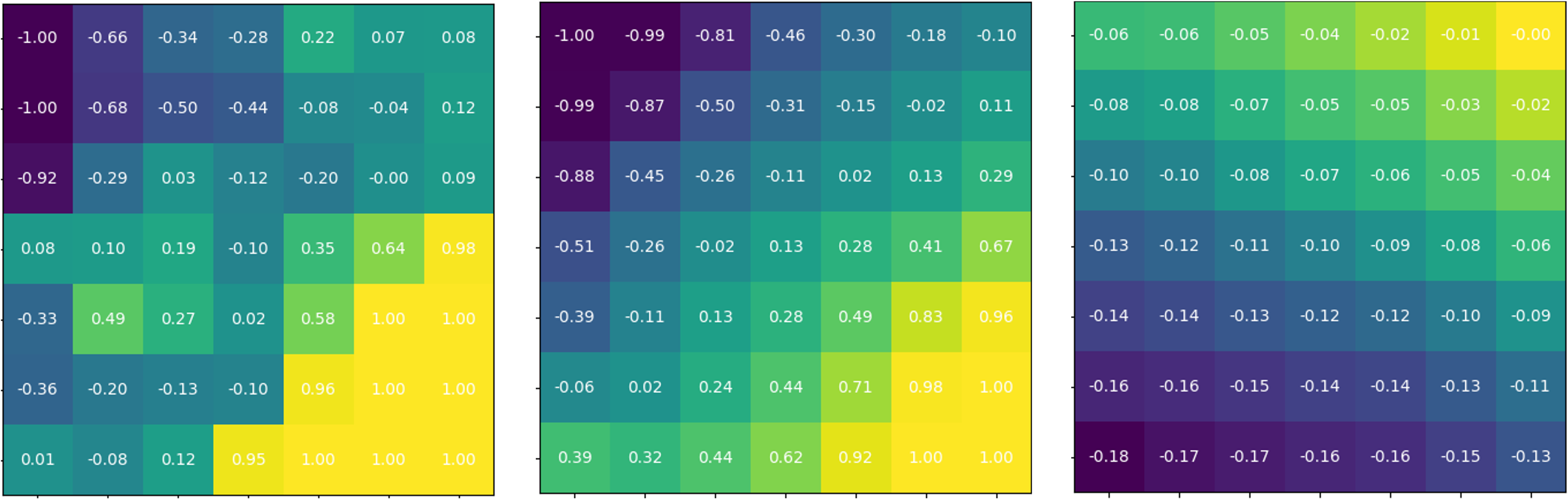}
    \caption{Variations in the initial reward values on 7x7 gridworld domain, when initialized using Xavier-Uniform across three different seeds. Numbers in each cell show the maximum reward,$\Rh$, that the agent upon taking an action in that cell.}
    \label{fig:uniform}
\end{figure*}

Even with uniform initialization methods \cite{orthonormal, kaiming}, we note that the initialization remains uniform in the parameter space $\psi \sim \Psi$ of the reward model $\Rpsi$, however, the subsequent reward over the domain of the state space under consideration is not at all uniform. In fact, as shown in figure \ref{fig:15x15_grid} (left), the initial reward values over a 15x15 gridworld space are ``patchy" or has certain ``peak" state locations and the rewards smoothen out from there. At the very least, such initial rewards heavily bias the sampled trajectories to query the human in the loop whereas in many of the initial queries the given two trajectories are extremely similar contributing to the cognitive load on the human in the loop. In fact, such initialization is assumed to be a ``feature than a bug" where works consider the initial randomness as a means for the agent to explore in the world. Although we agree that random initial rewards can in fact aid with coming up with diverse trajectories to query to the human in the loop, in reality, it is the reward parameters that are uniform and not the rewards themselves. In fact, we propose that the initialization of reward model parameters such that the corresponding reward values on the domain of the state space of the world are uniform will help with reward learning as well as stabilize the performance across various random seeds.

\begin{figure*}[h]
    \centering
    \includegraphics[width=0.8\textwidth]{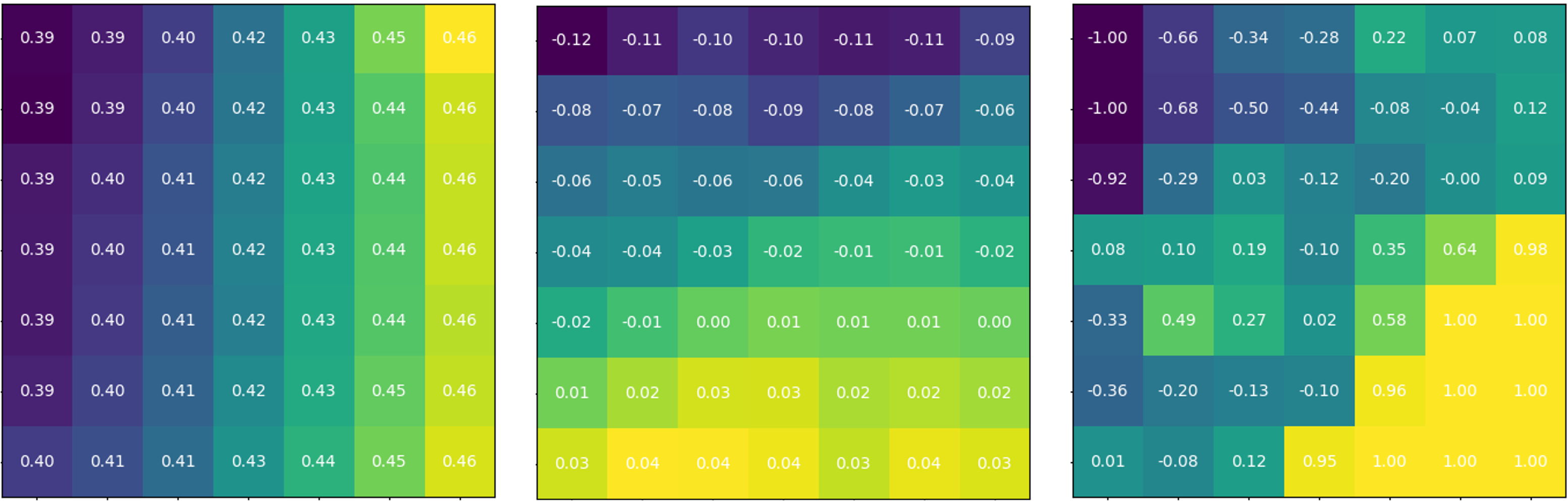}
    \caption{Variations in initial reward values on gridworld states as per Our data-driven approach (left), orthonormal initialization (middle), and Xavier initialization (right). Numbers in each cell show the maximum reward,$\Rh$, that the agent upon taking an action in that cell.}
    \label{fig:init}
\end{figure*}

\begin{figure}[h]
    \centering
    \includegraphics[width=0.45\textwidth]{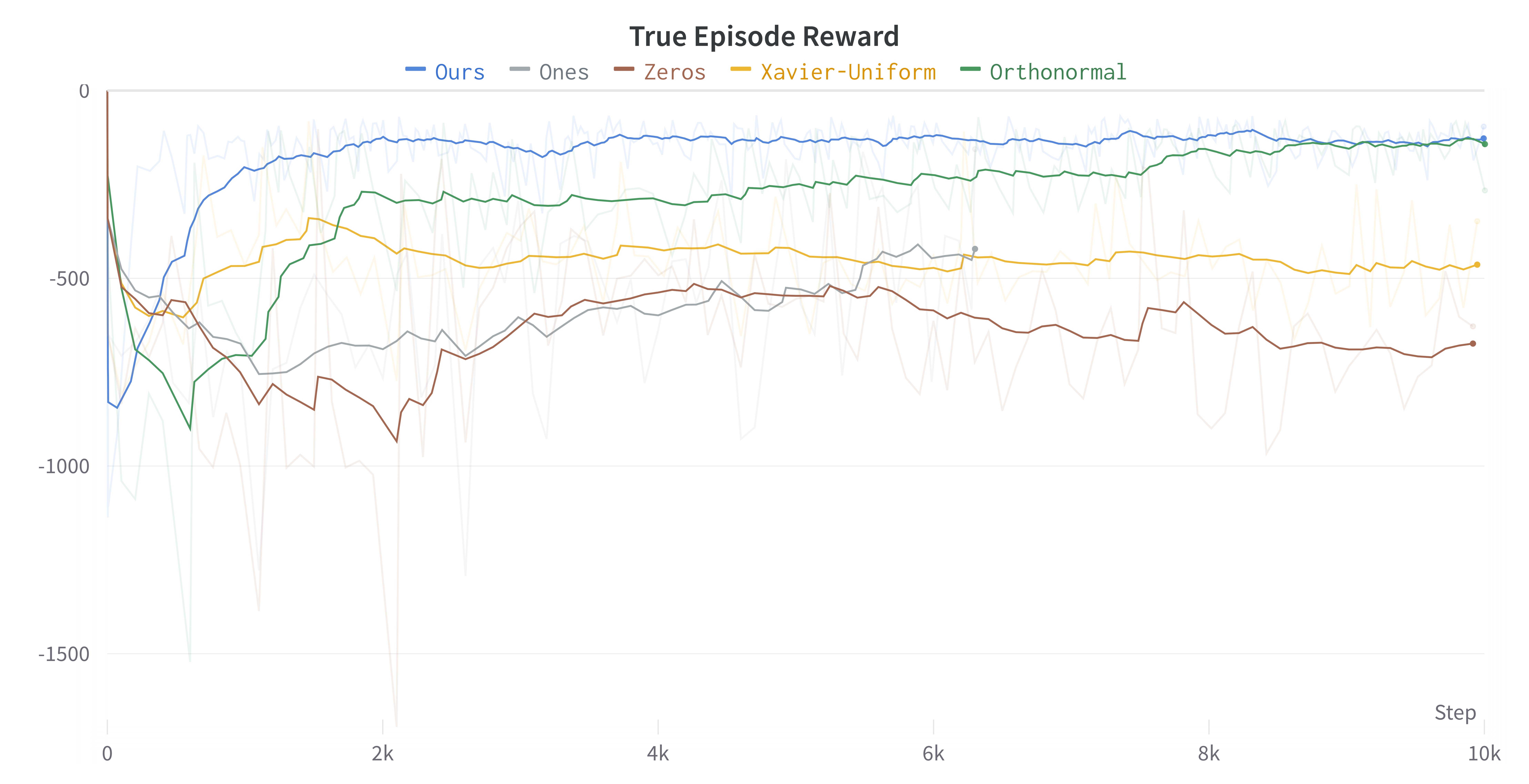}
    \caption{Return of the learned policy on the 7x7 gridworld as measured on the ground truth human reward $\Rh$ comparing Ours with baseline initializations.}
    \label{fig:7x7_results}
    \vspace{-0.1cm}
\end{figure}

\begin{figure}[h]
    \centering
    \includegraphics[width=0.45\textwidth]{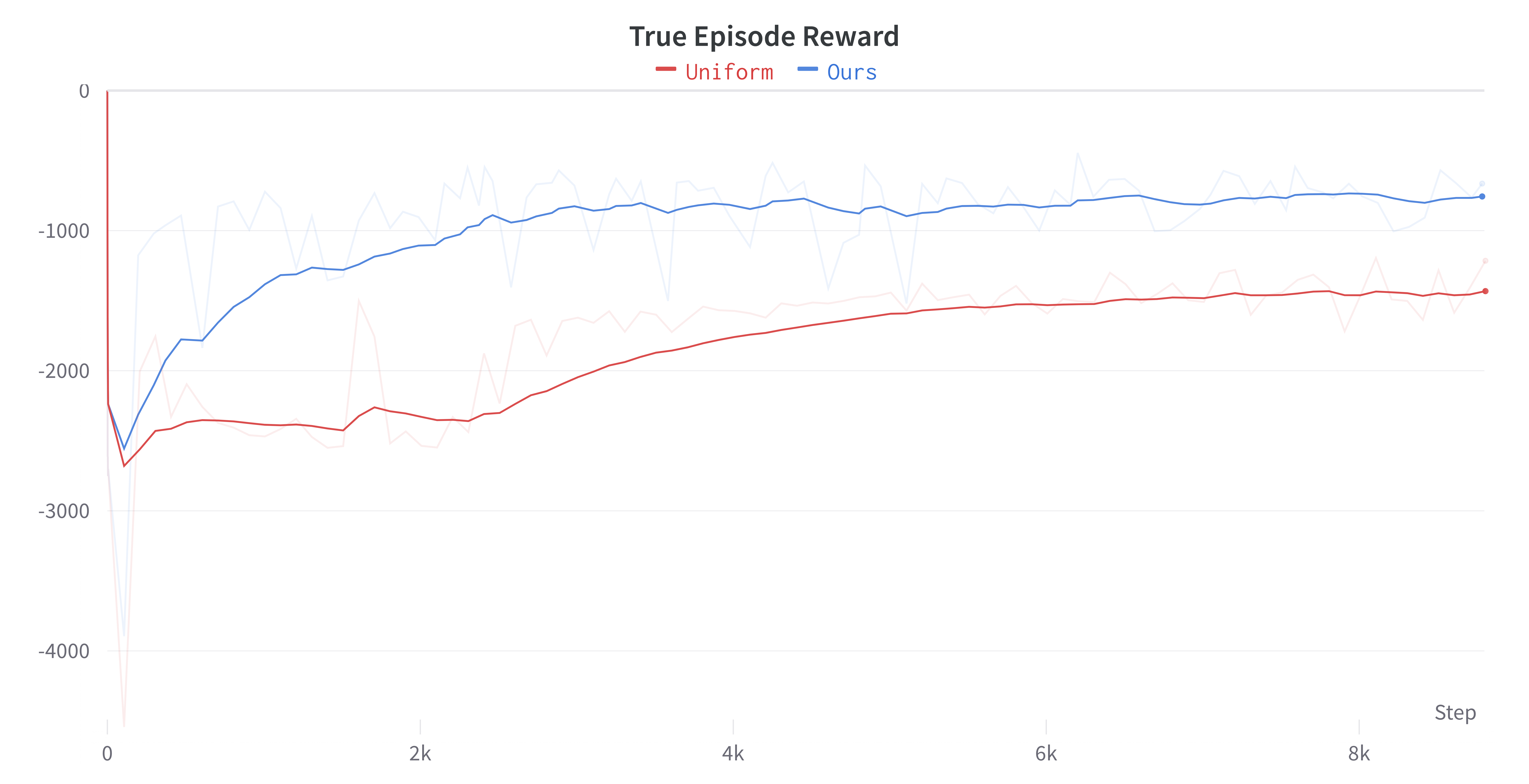}
    \caption{Return of the learned policy on the 15x15 gridworld as measured on the ground truth human reward $\Rh$.}
    \label{fig:15x15_results}
    \vspace{-0.1cm}
\end{figure}


\section{Method}
\vspace{-0.1cm}
To solve the issue of sensitivity of performance to reward initialization, we use our insight as mentioned in section \ref{sec:sensitivity} to propose our data-driven reward initialization method which ensures a uniform co-domain of the reward function $\hat{r}(s,a)$. Several PbRL works have utilized an unsupervised pre-training step that enables the agent to collect several trajectories in a buffer, from which trajectory pairs are sampled and queried to the user when the main training episodes begin. Say, the agent populates a buffer $D_\tau$ with trajectories $\tau_i$ by exploring the world via a policy $\pi$. We propose that ensuring that the predicted rewards over all the states visited in all the trajectories $s_i \in \tau_j \in D_\tau \; \forall i, j$ are the same, say some non zero value helps the agent learned much faster and ensures that the performance of the learned policy is consistent across various experimental runs (by varying random seeds). Hence, for the acquired dataset $D_\tau$, and some non-zero constant $\epsilon$, we solve a regression problem by minimizing the following pretraining-initialization loss, 
\begin{equation}
    \mathcal{L}_{P} = \sum_{\tau_j \sim D_\tau}\sum_{i=1}^{i=|\tau_j|} \norm{\Rpsi(s_i) - \epsilon}_2
\end{equation}

Note that $\Rpsi$ is updated using the pretraining-initialization loss only once to obtain a uniform $\Rpsi$. After this data-driven initialization, we follow the same algorithm as \cite{pebble} as our baseline. This additional update to the reward model has zero cost to the human in the loop and a one-time fixed cost for the agent (which is negligible in the context of training both $\Rpsi$ and $\pi$ in a PbRL algorithm.)

\section{Results}

To validate our proposed pretraining-initialization reward model loss, we conduct our experiments on a 7x7 gridworld and a 15x15 gridworld. The gridworld environment is square $n \times n$ matrix-like domain where the agent starts at the top left corner. The agent can take discrete action steps, such as, Up, Down, Left, and Right to navigate around the gridworld space. The underlying human reward, operationalized via oracular rewards, is assumed such that the goal location is the bottom right corner. We use the negative of euclidean distance from the current agent's location to the goal location as the underlying human reward function. This oracular reward, $\Rh$, is used to provide binary feedbacks when trajectory pairs are queried as: 

\begin{equation}
\label{eq:oracle_label}
    y(\tau_0, \tau_1) = \begin{cases}
    0 & \sum_{i} \Rh(\tau_0) > \sum_{i} \Rh(\tau_1) \\
    1 & \sum_{i} \Rh(\tau_0) < \sum_{i} \Rh(\tau_1) \\
    \end{cases}
\end{equation}

Figure \ref{fig:uniform} shows the high degree of variability in the initialized reward values for the gridworld states when Xavier \cite{xavier} weight initialized is used for the reward model. As indicated in previous sections, the predicted rewards are not uniform over the state locations and in fact vary quite significantly across different random seeds. Variability in initial reward estimates is not a problem in itself as the training process is expected to correct for it, however, in practice, we found that in cases when the reward function was initialized ``randomly" to align with the goal as in figure \ref{fig:uniform} (middle) the agent's performance was significantly better than when the initial reward estimates were incorrect, say, figure \ref{fig:uniform} (right).

We studied the impact of various popular initialization schemes for deep neural networks such as Kaiming Uniform \cite{kaiming}, Xavier Uniform \cite{xavier}, Orthonormal \cite{orthonormal}, Zeros and Ones. While initial rewards for the case of zeros weights and ones weight initialization is easy to visualize, figure \ref{fig:init} shows the reward initialization for Xavier (right), Orthonormal (middle), and Ours (left). It can be instantly noticed that in our data-driven initialization, with $\epsilon = 0.4$, almost all the states have a predicted reward value of $0.4$ which is as expected. However, for Xavier and Orthonormal initialization, the predicted rewards vary significantly across different states (hence, are ``patchy"). We allow 15 episodes of unsupervised pretraining steps for collecting trajectories in the PEBBLE backbone, which doubles as our dataset for minimizing $\mathcal{L}_P$ loss.

Figure \ref{fig:7x7_results} shows the results using various initialization strategies, averaged over 3 independent runs, (Ours, Ones, Zeros, Xavier-Uniform, Orthonormal) on a 7x7 gridworld domain with PEBBLE as the backbone PbRL algorithm implemented with DQN \cite{dqn} as the RL agent (in contrast to SAC \cite{haarnoja2018soft} which works with continuous action spaces), with all hyperparameters being the same across all runs. As expected, the data-driven reward initialization technique proposed in this work has very less variance compared to other initialization techniques (except zeros, where all the predicted rewards are also zeros, but it suffers from poor performance). Additionally, as discussed before, a uniform reward initialization over the state space improves exploration and allows the agent to collect more diverse trajectories thereby improving the queries made to the human in the loop. This also improves the overall performance of Our initialization technique. 

We also compare our data-driven initialization with Kaiming-Uniform initialization \cite{kaiming} on a 15x15 gridworld with similar semantics as our 7x7 gridworld. Figure \ref{fig:15x15_results} shows that even in the larger gridworld space our method can perform better than the baseline with zero additional cost to the human in the loop as well as a negligible cost to the PbRL agent. Figure \ref{fig:15x15_grid} shows the initial reward values as computed by both the methods Ours (right) and Kaiming-Uniform (left), and further highlights the issue of non-uniform reward values over the state space locations. On the other hand, even though (fig. \ref{fig:15x15_grid} (right)) reward initialization is not truly uniform, it is more consistent for unseen states (which are farther from the start location of the top-left corner). 


\section{Discussion}

In this work, we present the issue of the degeneracy of the reward model in preference-based reinforcement learning works, which in part is impacted by non-uniform predicted rewards over the state space domain. This can cause high variability in the performance measures across various runs and is found to be very sensitive to random seeds. We mitigate this issue by a data-driven reward initialization method that utilizes the states collected while performing unsupervised exploration (a common theme in several PbRL works) to ensure that the predicted rewards over all the states are the same by solving a regression problem. We validate our insights over two gridworld domains of sizes 7x7 and 15x15 which were chosen to visualize the predicted rewards.

We plan to further investigate the impact of reward initialization across more complex domains used in PbRL literature like locomotion tasks, robotic manipulation tasks, and other explicit discrete action domains. Additionally, this work calls for a more thorough investigation of potential reasons that worsen the issue of reward function degeneracy. Finally, we also plan to study the impact of reward initialization across various PbRL methods like \cite{surf, prior, christiano}.

\section*{Acknowledgements}
Kambhampati's research is supported by the J.P. Morgan Faculty Research Award, ONR grants N00014-16-1-2892, N00014-18-1-2442, N00014-18-1-2840, N00014-9-1-2119, AFOSR grant FA9550-18-1-0067 and DARPA SAIL-ON grant W911NF19-2-0006. 

\bibliography{aaai23} 
\end{document}